\documentclass[num-refs]{wiley-article}

\usepackage{siunitx}

\usepackage{rotating}
\usepackage{booktabs}
\usepackage{longtable} 
\usepackage{lscape} 
\usepackage{adjustbox}
\usepackage{algorithm}%
\usepackage{algorithmicx}%
\usepackage{algpseudocode}%

\usepackage{fancyhdr}
\usepackage{xcolor}

\newcolumntype{C}[1]{>{\centering\let\newline\\\arraybackslash\hspace{0pt}}m{#1}}
\papertype{Research Article}

\title{Adaptive transfer learning for surgical tool presence detection in laparoscopic videos through gradual freezing fine-tuning}

\author[1]{Ana Davila}
\author[2]{Jacinto Colan}
\author[1]{Yasuhisa Hasegawa}

\affil[1]{Institutes of Innovation for Future Society, Nagoya University, Nagoya, Aichi, 464-8601, Japan}
\affil[2]{Department of Micro-Nano Mechanical Science and Engineering, Nagoya University, Nagoya, Aichi, 464-8603, Japan}

\corraddress{Ana Davila, Institutes of Innovation for Future Society, Nagoya University, Nagoya, Aichi, 464-8601, Japan}
\corremail{davila.ana@robo.mein.nagoya-u.ac.jp}

\fundinginfo{Japan Science and Technology Agency (JST) CREST, Grant/Award Number: JPMJCR20D5. Japan Society for the Promotion of Science (JSPS) Grants-in-Aid for Scientific Research (KAKENHI), Grant/Award Number: 25K21247.

\hfill \break  \textbf{Ethics statement} 
 
 There are no human subjects or animal experiments in this article and informed consent is not applicable 
 
\hfill \break  \textbf{Data Availability Statement} 
 
The datasets used are publicly available at http://camma.u-strasbg.fr/datasets/ and https://ieee-dataport.org/open-access/cataracts. Our trained models are available at https://github.com/davilac/GFz-Tool.

\hfill \break \textbf{Conflict of Interest} 
 
 The authors declare no conflicts of interest. 
}

\runningauthor{Davila et al.}

\fancypagestyle{firstpage}{
  \fancyhf{}
  
  \fancyhead[L]{
    \vspace*{-25pt} 
    \parbox{\textwidth}{
      \textcolor{blue}{\small This article has been published in \textit{International Journal of Imaging Systems and Technology}, Vol. 35, no. 6, p. 70218  (2025). The final authenticated version is available online at: \url{https://doi.org/10.1002/ima.70218}}
    }
  }
}

\begin{document}

\begin{frontmatter}
\maketitle
\begin{abstract}
Minimally invasive surgery can benefit significantly from automated surgical tool detection, enabling advanced analysis and assistance. However, the limited availability of annotated data in surgical settings poses a challenge for training robust deep learning models. This paper introduces a novel staged adaptive fine-tuning approach consisting of two steps: a linear probing stage to condition additional classification layers on a pre-trained CNN-based architecture and a gradual freezing stage to dynamically reduce the fine-tunable layers, aiming to regulate adaptation to the surgical domain. This strategy reduces network complexity and improves efficiency, requiring only a single training loop and eliminating the need for multiple iterations. We validated our method on the Cholec80 dataset, employing CNN architectures (ResNet-50 and DenseNet-121) pre-trained on ImageNet for detecting surgical tools in cholecystectomy endoscopic videos. Our results demonstrate that our method improves detection performance compared to existing approaches and established fine-tuning techniques, achieving a mean average precision (mAP) of 96.4\%. To assess its broader applicability, the generalizability of the fine-tuning strategy was further confirmed on the CATARACTS dataset, a distinct domain of minimally invasive ophthalmic surgery. These findings suggest that gradual freezing fine-tuning is a promising technique for improving tool presence detection in diverse surgical procedures and may have broader applications in general image classification tasks.
\newline

\keywords{fine-tuning, transfer learning, surgical tool classification, minimally invasive surgery, gradual freezing, medical image analysis}
\end{abstract}
\end{frontmatter}

\section{Introduction}
\label{sec:1}

With advancements in video technology and information systems, future operating rooms are expected to incorporate computer-assisted tools to enhance surgeon decision-making and support context-aware systems, aiming to improve surgical outcomes and practices. A key component of these systems is surgical tool presence detection, also known as tool recognition, which verifies the presence of specific surgical instruments in every frame of a surgical video \cite{twinanda17endonet}. Analyzing surgical tool usage can assist in identifying surgical phases and actions \cite{padoy19machine, yamada24multimodal}, support autonomous robotic assistance \cite{fozilov23endoscope, yamada23task}, facilitate the creation of automatic reports, and aid in the evaluation of surgical skills \cite{jin18tool}.

Existing methods for tool recognition include sensor-based technologies and image-based approaches. Sensor-based technologies rely on embedded sensors to recognize or track surgical tools, with various types of sensors proposed, such as electromagnetic \cite{lahanas16simple}, optical \cite{elfring10assessment} and radiofrequency sensors \cite{kranzfelder13realtime}. Although effective, the use of sensors can be cumbersome, expensive, and difficult to embed in surgical settings. In contrast, image-based methods are more flexible and cost-effective, utilizing images from laparoscopic cameras. However, these methods face significant challenges related to image quality and loss of visibility \cite{ahmed24deep}. Surgical scenes are often occluded by blood or gas emissions from tools, complicating clear imaging. Additionally, the numerous potential combinations of tool usage make classification a complex multi-object detection task.

Recent progress in medical imaging tasks has been driven by the availability of large-scale, annotated medical image datasets and advances in deep learning, particularly Convolutional Neural Networks (CNNs). CNNs have demonstrated significant success in different classification tasks, including disease detection and abnormality classification. However, obtaining large amounts of labeled medical data can be challenging due to privacy concerns and patient data ownership \cite{kim22transfer}. Transfer learning offers a promising solution by reusing models pre-trained on extensive datasets such as ImageNet and adapting them to specific medical imaging tasks \cite{davila24comparison, raghu19transfusion, kora22transfer}. This approach not only mitigates the data scarcity issue but also reduces the computational resources and time required for training \cite{zhuang21comprehensive}. Several studies have demonstrated promising results by leveraging transfer learning for surgical tool presence detection tasks, formulating the surgical tool recognition problem as an image classification problem \cite{twinanda17endonet, vardazaryan18weakly, jin18tool, nwoye19weakly} or as an intermediate step for a higher-level task such as surgical phase or action recognition \cite{nwoye22rendezvous, kiyasseh23vision}. However, a significant issue for transfer learning applications is the domain differences between the pre-trained model source data and target images. These differences can lead to negative transfer, where the model performance deteriorates compared to training from scratch \cite{chen19catastrophic}. Additionally, catastrophic forgetting can occur when pre-trained knowledge is overwritten during adaptation to the target domain, resulting in the loss of previously learned information and limiting overall performance \cite{kirkpatrick17overcoming}. Addressing these challenges requires careful consideration of fine-tuning strategies and domain adaptation techniques to ensure effective transfer learning in surgical environments.

This paper introduces a novel fine-tuning strategy, Gradual Freezing, that focuses on the temporal dynamics of the training process to enhance knowledge transfer. Our approach is different from conventional fine-tuning methods that rely on static layer selection. Instead, our method dynamically modulates model plasticity during training. It begins by allowing the entire network to adapt and then progressively freezes layers based on their gradient-based stability. This staged reduction in trainable parameters acts as a form of adaptive structural regularization, creating a more stable optimization trajectory and preventing catastrophic forgetting. This process effectively navigates the critical trade-off between preserving general features and adapting to the surgical domain, a key challenge given the significant domain shift between natural images and surgical videos.

The main contributions of this work are:

\begin{itemize} 
\item A novel adaptive fine-tuning strategy based on gradual freezing that dynamically modulates model plasticity for a more stable and effective transfer learning process.
\item Demonstration of the proposed approach’s effectiveness for tool presence detection in laparoscopic videos using the Cholec80 dataset.
\item Benchmarking against state-of-the-art surgical tool presence detection and a wide range of fine-tuning based transfer learning approaches, establishing a new benchmark for surgical tool classification.
\item Demonstration of the fine-tuning strategy's broader applicability by testing its generalizability on the non-laparoscopic CATARACTS dataset.
\item Comprehensive ablation studies evaluating key parameters in our gradient-driven regularization approach and their impact on classification performance.
\end{itemize}

\section{Related works}
\label{sec:2}

Image-based surgical tool presence detection leverages video data analysis through various approaches. These can be broadly categorized by their use of temporal information \cite{wang24efficient} (frame-dependent vs. frame-independent) and multitasking capabilities \cite{tao23last} (single-task vs. multi-task learning). Relevant works are summarized in Table~\ref{tab:1}.

\begin{table}[bt]
\caption{Summary of existing deep-learning methods for surgical tool classification}\label{tab:1}
\begin{threeparttable}
\centering
\renewcommand{\arraystretch}{1.4}  
\scalebox{0.9}{
\begin{tabular}{l c c c c c}
\headrow
\bf{} & \bf{VFE backbone$^1$} & \bf{Frame dependence$^1$} & \bf{Multi-task} & \bf{Pretrained dataset} & \bf{Additional task} \\
Mishra et al. \cite{mishra17learning}   & ResNet50       & LSTM       & No     & ImageNet   & None \\
Alhali et al. \cite{alhajj18monitoring} & VGG-16         & LSTM       & No     & ImageNet   & None  \\
ConvLSTM \cite{nwoye19weakly}           & ResNet18       & LSTM       & No     & ImageNet   & None  \\ 
Abdulbaki et al. \cite{abdulbaki21deep} & ResNet50       & LSTM       & No     & ImageNet   & None  \\
LapTool-Net \cite{namazi22contextual}   & Inception v1   & GRU        & No     & None       & None \\
Shi et al. \cite{shi20realtime}         & VGG-16         & Attention  & Yes    & ImageNet   & Localization  \\
ToolNet\cite{kondo21lapformer}          & ResNet50       & Attention  & No     & ImageNet   & None  \\
LAST \cite{tao23last}                   & Swin-B         & Attention  & Yes    & ImageNet   & Phase  \\
Endo3D \cite{chen18endo3d}              & CNN            & LSTM       & Yes    & sport1M    & Phase  \\
Wang et al. \cite{wang19graph}          & DenseNet-121   & GCN        & No     & ImageNet   & None  \\
Wang et al. \cite{wang24efficient}      & Swin-B         & HMM        & Yes    & ImageNet   & Phase  \\
ToolNet \cite{twinanda17endonet}        & AlexNet        & None       & No     & ImageNet   & None  \\ 
Sahu et al. \cite{sahu16tool}           & AlexNet        & None       & No     & ImageNet   & None  \\ 
Jaafari et al. \cite{jaafari21towards}  & Inception ResNet v2 & None  & No     & ImageNet   & None   \\
EndoNet\cite{twinanda17endonet}         & AlexNet        & HMM        & Yes    & ImageNet   & Phase  \\ 
Kanakatte et al. \cite{kanakatte20surgical} & ResNet     & LSTM       & Yes    & ImageNet   & Segmentation   \\ 
FR-CNN \cite{jin18tool}                 & VGG-16         & None       & Yes    & ImageNet   & Localization   \\ 
MTRC-Net\cite{jin20multitask}           & ResNet50       & LSTM       & Yes    & ImageNet   & Phase  \\ 
Vardazaryan et al. \cite{vardazaryan18weakly} & ResNet18 & None       & Yes    & ImageNet   & Localization  \\ 
Jalal et al. \cite{jalal23laparoscopic} & ResNet50       & LSTM       & Yes    & ImageNet   & Localization \\
\hline
\multicolumn{6}{l}{ $^1$ For the cases in which several architectures are proposed, only the best performing is reported here.} \\
\end{tabular}}
\end{threeparttable}
\end{table}

Frame-dependent methods analyze sequential frames to capture tool motion and interactions. Architectures based on Recurrent Neural Networks (RNNs) or Attention-based networks are commonly used to exploit the temporal dependencies between frames. Each video frame usually follows a visual feature extraction (VFE) module, commonly based on CNN backbones, with their outputs fed to a second architecture for modeling temporal relationships. Mishra et al. \cite{mishra17learning} proposed a stacked LSTM-based network to extract temporal dependencies from sequences of visual features extracted from a CNN network. Al Hajj et al. \cite{alhajj18monitoring} introduced a boosting strategy for training a CNN-RNN architecture, exploiting the RNN’s output to supervise CNN training and combine visual features with temporal context. Nwoye et al. \cite{nwoye19weakly} proposed a CNN architecture enhanced with convolutional LSTMs called ConvLSTM, weakly supervised on tool binary annotations for presence detection and tracking. Abdulbaki et al. \cite{abdulbaki21deep} presented a hierarchical architecture with a CNN and two LSTM models, where the CNN learns spatial features and the LSTMs learn temporal dependencies from video data. Namazi et al. \cite{namazi22contextual} introduced LapTool-Net, a multilabel classifier that combines CNNs for spatial feature extraction and GRUs for temporal feature extraction to detect surgical tools in laparoscopic video frames, leveraging inter-frame tool usage and surgical step correlations.

Attention mechanisms have also been explored in several works due to their ability to model complex time dependencies \cite{shi20realtime, kondo21lapformer, tao23last, yin23error}. Shi et al. \cite{shi20realtime} developed a real-time attention-guided convolutional neural network (CNN) for frame-by-frame detection of surgical tools, which comprises coarse and refined detection modules. Kondo et al. \cite{kondo21lapformer} proposed LapFormer, combining a CNN visual feature extractor with attention-based temporal analysis. Tao et al. \cite{tao23last} introduced Latent Space-constrained Transformers (LAST) for automatic surgical phase recognition and tool presence detection, which comprises a Swin-B backbone for visual feature extraction and a transformer variational autoencoder (VAE) for learning semantic interactions between phases and tools. Yin et al. \cite{yin23error} propose integrating a 3D attention mechanism into the network backbone to improve the simultaneous classification and localization of surgical tools in the m2cai16-tool-location dataset.

Alternatively, some works have used different approaches for modeling temporal dependencies to reduce computational resources. Chen et al. \cite{chen18endo3d} used a 3D CNN for spatiotemporal feature learning from short clips. Wang et al. \cite{wang19graph} implemented Graph Convolutional Networks (GCNs) to learn temporal relationships between frames, achieving significant performance improvement by incorporating unlabeled frames alongside labeled ones. Wang et al. \cite{wang24efficient} used a hidden Markov model (HMM) stabilized deep learning method for surgical tool presence detection instead of computationally expensive LSTM or attention mechanisms.

While providing a more comprehensive understanding of tool usage, frame-dependent models can result in complex architectures that are computationally expensive to train. They also require the use of blocks of sequential data or buffers that feed the network at each inference request, with the size of this buffer impacting prediction performance greatly, thereby limiting its applications for online recognition. Frame-independent methods, on the other hand, treat each video frame as a standalone image, employing Convolutional Neural Networks (CNNs) for independent tool classification within each frame. ToolNet, introduced by Twinanda et al. \cite{twinanda17endonet}, is an early deep learning-based method for surgical tool detection, utilizing a pretrained AlexNet CNN backbone and fine-tuned for the Cholec80 dataset. Recognizing the imbalanced nature of surgical datasets, Sahu et al. \cite{sahu16tool} proposed creating balanced training datasets focusing on tool occurrences for a multi-label classification task, trained a CNN, and used post-processing temporal smoothing to incorporate previous frame detections, thus minimizing false detections without the high computational cost of temporal data during training. Jo et al. \cite{jo19robust} developed a real-time CNN-based algorithm for the detection of surgical instruments. The proposed algorithm uses the object detection system You Only Look Once (YOLO9000), incorporating motion vector prediction to detect missing surgical tools in successive frames. Jaafari et al. \cite{jaafari21towards} avoided the use of RNNs and addressed the unbalanced data problem in the Cholec80 laparoscopy video dataset using multiple data augmentation techniques and fine-tuning the InceptionResnet-v2 network pretrained on the ImageNet dataset for surgical tool classification.

Most of the previously mentioned approaches were single-task learning, focusing solely on the tool presence detection task, which provides a straightforward approach but potentially lacks contextual understanding. Multi-task learning addresses this by considering the inherent relationship between surgical tools and intraoperative information such as surgical phases or actions. This approach aims for a holistic understanding of the surgical workflow by using additional information like surgical phase labels during training. While offering comprehensive analysis, multi-task learning necessitates complex model architectures and additional data labeling efforts. Twinanda et al. \cite{twinanda17endonet} introduced Endonet, based on an AlexNet architecture, which integrates two independently trained architectures: ToolNet, for surgical tool recognition solely, and PhaseNet, for surgical phase prediction. Endonet achieves both phase recognition and surgical tool presence detection concurrently in cholecystectomy videos, surpassing the performance of single-prediction models. Kanakatte et al. \cite{kanakatte20surgical} combined frame-level ResNet features, video-level Inflated Inception 3D (I3D) features, and spatio-temporal LSTM layers for tool detection and segmentation. Jin et al. \cite{jin18tool} proposed a tool detection method using Faster R-CNN with a VGG-16 base architecture, showing effectiveness in detecting spatial bounds and tool presence in the m2cai16-tool dataset. In a subsequent work, Jin et al. \cite{jin20multitask} introduced a multi-task deep learning framework enhanced by a correlation loss exploiting the tool-phase relationship. Vardazaryan et al. \cite{vardazaryan18weakly} utilized a deep learning architecture for simultaneous tool presence detection and localization, incorporating convolutional layers into a ResNet-10 base model. Jalal et al. \cite{jalal23laparoscopic} proposed a weakly-supervised deep learning approach, using a ResNet-50 with attention modules and feature fusion for multi-task classification: surgical phase recognition, tool classification, and tool localization. This approach integrates LSTM for temporal information modeling, achieving state-of-the-art performance on the Cholec80 dataset.

\subsection{Transfer learning}
The limited availability of annotated surgical videos presents a significant challenge for training supervised deep learning models for surgical tool detection. To address this, researchers often leverage pre-trained models on large, publicly available datasets such as ImageNet (as shown in Table \ref{tab:1}). These pre-trained models typically capture generic visual features that can be beneficial for tasks in related domains. A widely utilized method for leveraging pre-trained models is full fine-tuning, where the classifier layer of the pre-trained model is replaced with a randomly initialized one conditioned to fit the target classes. The model weights are initialized with the pre-trained values, and then all layers are further trained on the target surgical video dataset \cite{hinton06reducing}. Although this remains the primary method for transfer learning, challenges persist due to domain differences between the source dataset and the target surgical domain. These differences can lead to negative transfer, where the pre-trained model hinders performance compared to training from scratch \cite{chen19catastrophic}. Furthermore, catastrophic forgetting can occur, where the model’s pre-trained knowledge is overwritten by significant gradient flows from the new layer to earlier layers during adaptation to the surgical domain, hindering its ability to leverage previously learned information and ultimately limiting its overall performance \cite{kirkpatrick17overcoming}. Furthermore, a significant challenge in endoscopic images is the natural shift, which refers to distribution shifts arising from variations such as camera angles, lighting conditions, and tool appearances \cite{taori20measuring}. 

To preserve information from the pre-trained model and reduce overfitting during fine-tuning, several approaches have proposed freezing specific parameters in the pre-trained model \cite{royer20flexible, kumar22fine}. 
A common practice is to freeze the initial layers and only fine-tune the latter layers. This method preserves the general features learned from the source domain while adapting the model to the target task \cite{yosinski14howtransferable}. Other approaches propose starting with all model layers frozen and gradually unfreezing them to reduce the possibility of catastrophic forgetting \cite{mukherjee20distilling, howard18universal}. However, these approaches may not be optimal for all distribution shifts, as different layers may be more relevant or sensitive to specific types of data change \cite{guo19spottune}. Kumar et al. \cite{kumar22fine} demonstrated that while full fine-tuning can lead to better accuracy within distributions similar to the source domain, it may underperform for out-of-distribution data. In contrast, linear probing followed by full fine-tuning, which involves initially updating only the last linear layer of a pre-trained model and then fine-tuning the entire network, might perform better for out-of-distribution data, but could lead to reduced accuracy for in-distribution data. Additionally, several studies have investigated the problem of automatically selecting the parameters to freeze, exploring the relationship between the subset of layers to freeze and the distribution shift between the source and target domains \cite{lee23surgical, liu21autofreeze}. 

Recent research on transfer learning has extensively explored the role of learning rate and regularization in preventing overfitting and facilitating the retention of valuable knowledge from pre-trained models for adaptation to the target task \cite{lee23surgical, kumar22fine, shen21partial, ro21autolr, li20rethinking, li18explicit}. Shen et al. \cite{shen21partial} introduced a general method in which they freeze or fine-tune specific layers based on evolutionary search algorithms to efficiently determine optimal layers and learning rates for the target task. Ro et al. \cite{ro21autolr} presented an algorithm that adjusts learning rates and prunes the weights of each layer according to their role and importance. Li et al. \cite{li20rethinking} introduced a strategy of using a lower learning rate during fine-tuning to reduce the risk of losing previous knowledge acquired during pre-training. Recently, Lee et al. \cite{lee23surgical} proposed surgical fine-tuning, where the learning rate of blocks of layers is modified based on the relative gradient change at each training iteration. Larger weight gradients are assumed to be important for the new target task and therefore maintain a large learning rate, while low or negligible weight changes represent layers not relevant to the target task and therefore can be omitted from the training process. Similar strategies for selecting optimal layers have been proposed in medical image classification, with prior works also relying on selecting the best combination of layers to optimize knowledge transfer \cite{davila24comparison, vrbancic20transfer, mukhlif23incorporating, davila23gradient}. 

As an alternative to freezing parameters, parameter-efficient fine-tuning methods such as low rank Adaptation (LoRA) \cite{hu21lora} aim to reduce the number of trainable parameters. LoRA introduces low-rank matrices into the weight updates of each layer, significantly lowering trainable parameters while often maintaining comparable performance. This approach is particularly beneficial for large models used in transformer architectures, as it allows efficient adaptation without the computational demands of full fine-tuning, offering a balance between full fine-tuning and freezing-unfreezing strategies. Recent studies have applied LoRA and Mona adapters to vision-language models for medical ultrasound, achieving efficient knowledge transfer \cite{qu25adapting}.

Beyond parameter-efficient fine-tuning methods that add trainable components, another prominent direction for improving computational efficiency is model pruning. This family of techniques aims to remove redundant weights, channels, or even entire layers from a pre-trained network. Recent advancements have leveraged bio-inspired and metaheuristic optimization algorithms to automate this process. For example, Ismail et al. \cite{ismail24genetic} proposed using a Genetic Algorithm (GA) to find an optimal fine-tuning strategy for pre-trained models by selectively pruning or freezing layers. Similarly, Tmamna et al. \cite{tmamna24bpso} introduced BPSO-LPruner, which uses Binary Particle Swarm Optimization (BPSO) for layer pruning. Taking this concept further, BioTune \cite{davila25biotune}, uses evolutionary optimization to not only select which layers to freeze but also to adjust the learning rates of the remaining trainable layers. Although effective, these search-based methods treat the selection of layers as a complex combinatorial optimization problem. They can find highly optimized static configurations but often introduce significant computational overhead from the search process itself, and there is a risk of degrading model performance if not carefully tuned.


Our proposed approach optimizes the fine-tuning training procedure through an adaptive layer freezing strategy. This method differs from existing research by gradually freezing different subsets of layers based on their importance, which is determined by their weight gradients after each training iteration. This method mitigates catastrophic forgetting, preserves valuable learned features, and adapts the model to new tasks, without changing the final model architecture. As a result, it shows superior performance in detecting the presence of surgical tools compared to traditional fine-tuning strategies.

\section{Methodology}
\label{sec:3}

\begin{figure}[!htbp]	
\centering
\includegraphics[width=0.67\linewidth]{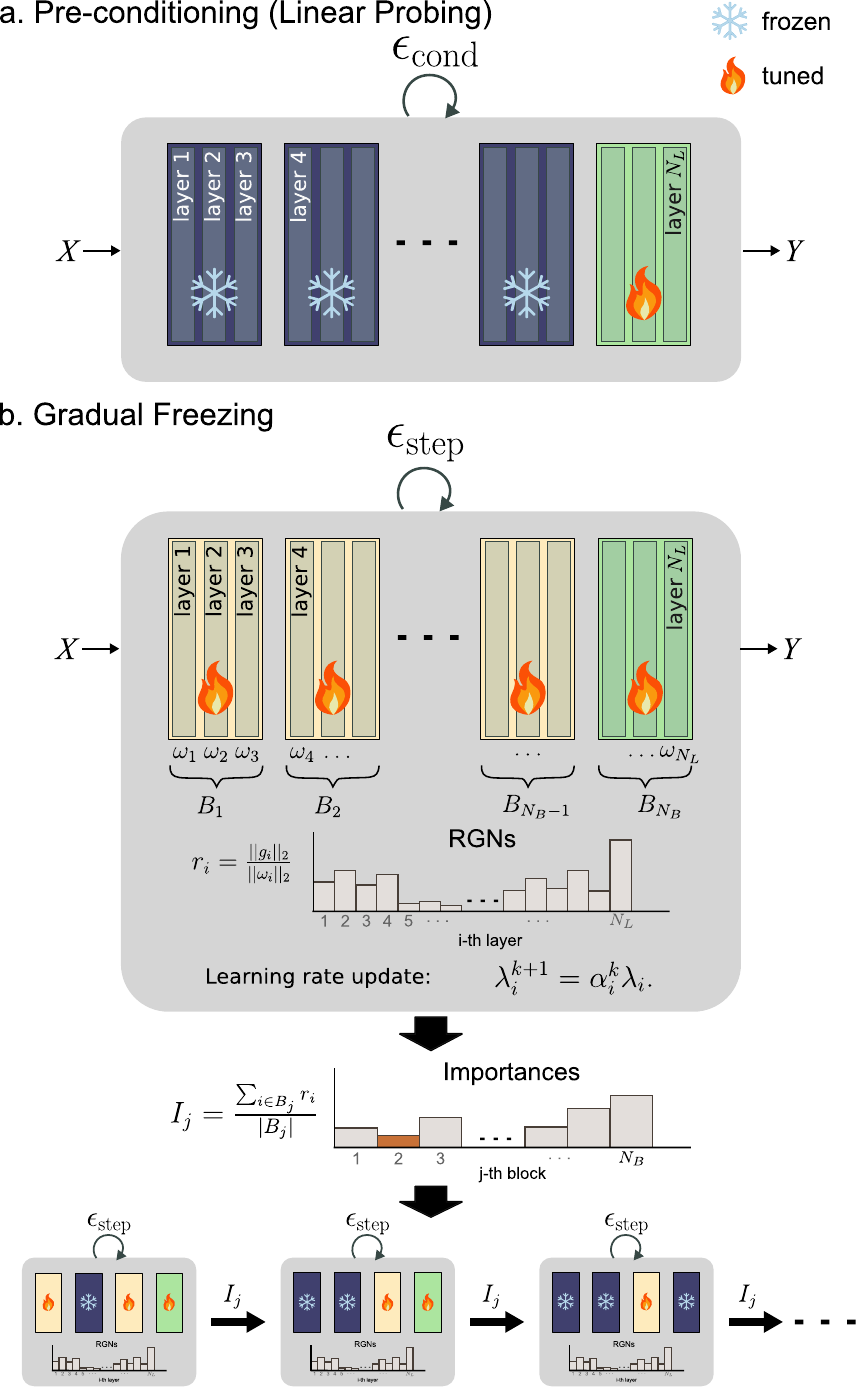}
\caption{Overview of the proposed gradual freezing fine-tuning strategy for adaptive transfer learning. \textbf{a. Pre-conditioning:} All layers except the last classifier layer are frozen with pre-trained parameters and trained for $\epsilon_{\text{cond}}$ epochs, allowing the new, randomly initialized layer to adapt to the new task (Linear probing). \textbf{b. Gradual freezing:} All layers are unfrozen and layers with low importance, based on their relative gradient norm (RGN), are gradually frozen at each subsequent freezing step (every $\epsilon_{\text{step}}$ epochs). \label{fig:1}} 
\end{figure}

This section presents our approach for adaptive fine-tuning based on gradual freezing. Our method utilizes a model pre-trained on a large, generic dataset to extract general visual features, which are then refined for the specific task of surgical tool detection. The workflow of the proposed fine-tuning algorithm is illustrated in Fig.~\ref{fig:1}, and Algorithm~\ref{alg:1} summarizes the overall procedure.

\begin{algorithm}[h]
\caption{Adaptive Fine-Tuning Algorithm}
\label{alg:1}
\begin{algorithmic}[1]
\Require $N_L$-layer pre-trained model $\mathcal{M}$, base learning rates $\lambda_i$, number of target classes $n_{out}$
\State {\bf Replace} last classifier layer $L_{N_L}$ with a new $n_{\text{out}}$-dimensional layer
\State {\bf Initialize} new layer $L_{N_L}$ with random weights and bias
\State {\bf Freeze} pre-trained weights: $\omega_1, \dots, \omega_{N_L-1}$
\For{it = 0 \textbf{to} $\epsilon_{\text{cond}}$} \Comment{// Pre-conditioning}
    \State {\bf Train} model
\EndFor
\State {\bf Unfreeze} all pre-trained weights
\While {no\_improvement\_epochs $<$ patience} \Comment{// Gradual Freezing}
\For{it = 0 \textbf{to} $\epsilon_{\text{step}}$}
        \State {\bf Train} model
        \State {\bf Record} relative gradient norms (RGNs) $r_i^k$ \quad (Eq.~\ref{eq.1})
        \State {\bf Update} learning rates based on $\alpha_i^k$ \quad (Eq.~\ref{eq.2}-\ref{eq.3})
    \EndFor
    \If {layers remaining to freeze $>$ 0}
        \State {\bf Compute} block importance index $I_j^k$ \quad (Eq.~\ref{eq.4})
        \State {\bf Rank} blocks based on their importance index
        \State {\bf Freeze} block with the lowest importance index
    \EndIf
\EndWhile
\end{algorithmic}
\end{algorithm}

\subsection{Network architecture and pre-training}
A CNN-based network architecture is utilized, consisting of $N_L$ layers. All model layers are grouped into $N_B$ blocks ($N_B \leq N_L$) based on their function (e.g., layers belonging to a residual block or a fully connected classifier) or an arbitrary criterion (such as a predefined percentage of total layers). The notation $B_j$ is used to denote the set of layers belonging to block $j$. The network layers are initialized with pre-trained weights $\omega_S^0$. For the fine-tuning process, a base learning rate $\lambda_i$ is established for each layer $i \in {1, \dots, N_L}$.

\subsection{Adaptive fine-tuning strategy}
Our approach follows a two-phase fine-tuning approach:

\subsubsection{Pre-conditioning Phase}
In the pre-conditioning phase, all layers of the pre-trained model, except for the last classifier layer, are frozen (i.e., their weights are kept constant and not updated during training). The classifier layer is replaced with a new, randomly initialized layer specifically designed to fit the target classes. This new layer is then trained for $\epsilon_{\text{cond}}$ epochs. This procedure, also known as linear probing \cite{kumar22fine}, serves to adapt the new classifier to the feature distribution of the target data. Since only the final layer is trained, this step avoids large gradient updates from propagating back through the network, which could otherwise negatively affect the pre-trained weights in the feature extraction layers. This preserves the generic features from the pre-trained model while the model's output layer adapts to the new task.

\subsubsection{Gradual Freezing Phase}
Once pre-conditioning is complete, the Gradual Freezing phase begins. The entire model, including the output layer, is unfrozen and retrained on the target dataset for $\epsilon_{\text{step}}$ epochs, following a full fine-tuning approach. During this stage, all layers adapt to the target domain collectively. At each epoch, the relative gradient norms (RGNs) are computed for each layer and normalized to dynamically update the learning rate of each layer. This step follows an auto-RGN approach \cite{lee23surgical}, where relative gradients are used as indicators for learning performance. This approach has been explored in various nonlinear optimization applications based on gradient based methods \cite{davila24realtime, colan24variable}. The RGN $r_i^k$ for layer $i$ at epoch $k$ is given by:

\begin{equation}
r_i^k = \frac{||g_i^k||_2}{||\omega_i^k||_2} \quad \text{for} \quad i \in \{1,\dots,N_L\},
\label{eq.1}
\end{equation}

where $g_i^k$ and $\omega_i^k$ correspond to the gradient and weights for layer $i$ at training iteration $k$. Using the RGN is technically advantageous because it provides a scale-invariant measure of change. The raw gradient norm $||g_i^k||_2$ is sensitive to the scale of a layer's weights; normalizing it by the weight norm $||w_i^k||_2$ quantifies the update relative to the layer's current parameter magnitudes, enabling a more stable and meaningful comparison across layers with different architectures and scales. This process can be viewed as an implicit, layer-level method for retaining relevant feature hierarchies, rather than an explicit selection of individual features in the traditional sense. The learning rate weight $\alpha_i^k \in [0,1]$ is defined as the normalized version of $r_i^k$ as:

\begin{equation}
\alpha_i^k = \frac{r_i^k}{r_{\text{max}}^k} \quad \text{for} \quad i \in \{1,\dots,N_L\},
\label{eq.2}
\end{equation}

where $r_{\text{max}}^k$ corresponds to the maximum RGN among all layers at a given epoch $k$. These weights are used to update learning rates for the subsequent epoch by:

\begin{equation}
\lambda_i^{k+1} = \alpha_i^k \lambda_i.
\label{eq.3}
\end{equation}

This update acts as a per-layer adaptive learning rate schedule. Layers with a high RGN (indicating high learning activity) receive a learning rate closer to the base rate $\lambda_i$, while layers with a low RGN (indicating stabilized layers) have their learning rates dampened. This per-layer control of the learning rate helps to preserve the knowledge in more stable layers by reducing the magnitude of their weight updates, while allowing more active layers to continue adapting to the new task. After $\epsilon_{\text{step}}$ training iterations, the groups of layers are ranked based on their importance index $I_j^k$ defined as:

\begin{equation}
I_j^k = \frac{\sum_{i \in B_j} r_i^k}{|B_j|} \quad \text{for} \quad j \in \{1,\dots,N_B\},
\label{eq.4}
\end{equation}

where $B_j$ represents the set of layers in block $j$ and $|B_j|$ is the number of layers in block $j$. This importance index is the average RGN for a block of layers. A high index indicates that the block's weights are undergoing larger changes in response to the target data, while a low index suggests the block's weights have stabilized. Layers belonging to the group with the lowest importance index are then frozen. The training loop continues for another $\epsilon_{\text{step}}$ epochs, after which the importance index is recomputed, and a new set of layers is frozen. This cycle repeats until there is no further improvement in validation accuracy, triggering an early stop. If only a single group of layers remains unfrozen, it is kept unfrozen until early stopping is activated. The definition of these layer groupings ($B_j$) is a key configurable aspect of the framework. Strategies can range from grouping by functional architectural blocks to selecting individual layers, allowing the method to be adapted to different network architectures and fine-tuning objectives.

The motivation for this gradual approach is to mitigate catastrophic forgetting. When all layers are unfrozen and trained simultaneously, the model risks altering the general features learned during pre-training. By identifying and freezing layers with low learning activity (low importance index), their learned feature representations are preserved. This method concentrates the training updates on the layers that are still adapting to the new task (high importance index). This is complemented by the dynamic per-layer learning rates, which adjust updates based on weight changes. A further outcome is improved training efficiency, as freezing layers reduces the number of parameters that require gradient computation and backpropagation in subsequent epochs.

\section{Experimental setup}
\label{sec:4}

\subsection{Datasets and training splits}
\begin{figure}[tb]	
\centering
\includegraphics[scale=0.5, width=1\linewidth]{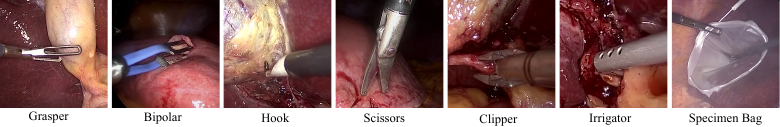}
\caption{Sample images of labeled surgical tools in the Cholec80 dataset. \label{fig:2}}
\end{figure}

We validate our proposed approach for surgical tool presence detection using the Cholec80 dataset \cite{twinanda17endonet}. This dataset comprises 80 endoscopic videos of cholecystectomy procedures, a laparoscopic surgery that involves surgical removal of the gallbladder, performed by 13 different surgeons. The videos have a resolution of 1920x1080 or 854x480, and are recorded at a frame rate of 25 frames per second. On average, the videos have a duration of 2095 seconds, with individual video lengths ranging from 739 s to 5993 s. For the purpose of annotation, the videos were downsampled to 1 frame per second. A tool is marked as in use if it is visible through the endoscope, specifically if at least half of the tool tip is within the frame. The annotations cover seven specific surgical tools: Grasper, Bipolar, Hook, Scissors, Clipper, Irrigator, and Specimen bag, illustrated in Fig.~\ref{fig:2}, and provide a binary label for each image and tool. The number of frames in which a tool appears is highly imbalanced, as shown in Table~\ref{tab:2}.

There are notable differences in the dataset splits used for reporting metric results in prior studies on surgical tool presence detection. Typically, models were trained using the first 40 videos from the Cholec80 dataset. However, the evaluation datasets varied across studies. Some studies used the last 40 videos for evaluation, here referred to as the L40 split \cite{twinanda17endonet, chen18endo3d, wang19graph, jin20multitask, abdulbaki21deep, jalal23laparoscopic}. Other studies opted for the last 30 videos for evaluation, which we will refer to as the L30 split \cite{vardazaryan18weakly, nwoye19weakly, abdulbaki21deep}. Additionally, in \cite{jaafari21towards}, the training was conducted with 60 videos, and the evaluation was performed on the remaining 20 videos, termed the L20 split. To ensure a fair comparison of our proposed approach with existing methods, we evaluated the performance using both the L40 and L30 splits. 

\begin{table}[bt]
\caption{Number of training frames where a tool appears in Cholec80 dataset}\label{tab:2}
\centering
\renewcommand{\arraystretch}{1.4}  
\scalebox{0.9}{
\begin{tabular}{l c c}
\headrow
\bf{Tool}       & \bf{Number of frames}   & \bf{Percentage (\%)} \\ 
Grasper         & 102615             & 55.59 \\ 
Bipolar         & 8876               & 4.81 \\ 
Hook            & 103106             & 55.86 \\ 
Scissor         & 3254               & 1.76 \\ 
Clipper         & 5986               & 3.24 \\ 
Irrigator       & 9814               & 5.32 \\ 
Specimen bag    & 11463              & 6.21 \\ 
Total           & 184578             & 100.00 \\
\hline
\end{tabular}}
\end{table}

To evaluate the generalizability of our approach, we conducted additional experiments on the CATARACTS dataset \cite{alhajj21cataracts}. This dataset contains videos of cataract surgeries, a different surgical domain from the cholecystectomy procedures in Cholec80. It features different anatomy, lighting conditions, and a distinct set of surgical tools. For our analysis, we downsampled these videos to 1 frame per second. We followed the standard data splits and utilized the provided multi-label tool presence annotations, focusing on classifying the surgical tools detailed in Table~\ref{tab:3}.

\begin{table}[bt]
\caption{Number of frames where a tool appears in CATARACTS dataset}\label{tab:3}
\centering
\renewcommand{\arraystretch}{1.4}  
\scalebox{0.9}{
\begin{tabular}{l c c}
\headrow
\bf{Tool}                       & \bf{Number of frames}   & \bf{Percentage (\%)} \\ 
Charleux cannula                & 416   &  1.28\\ 
hydrodissection cannula         & 674   &  2.08\\ 
Rycroft cannula                 & 960   &  2.96\\ 
viscoelastic cannula            & 760   &  2.35\\ 
cotton                          & 128   &  0.40\\ 
capsulorhexis cystotome         & 1519  &  4.69\\ 
Bonn forceps                    & 321   &  0.99\\ 
capsulorhexis forceps           & 406   &  1.25\\
Troutman forceps                & 94    &  0.29\\
needle holder                   & 41    &  0.13\\
irrigation/aspiration handpiece & 5066  &  15.65\\
phacoemulsifier handpiece       & 5197  &  16.05\\
vitrectomy handpiece            & 579   &  1.79\\
implant injector                & 499   &  1.54\\
primary incision knife          & 244   &  0.75\\
secondary incision knife        & 176   &  0.54\\
micromanipulator                & 5870  &  18.13\\
suture needle                   & 86    &  0.27\\
Mendez ring                     & 31    &  0.10\\
Vannas scissors                 & 41    &  0.13\\
Total                           & 32380 & 100.00 \\
\hline
\end{tabular}}
\end{table}

\subsection{Pre-trained model architectures}
We implement our approach using two pre-trained CNN architectures known for their strong performance in image recognition: ResNet-50 \cite{he16deep} and DenseNet-121 \cite{huang17densely}. ResNet-50, a 50-layer CNN, utilizes shortcut connections to mitigate the vanishing gradient problem and contains 25.6 million parameters. DenseNet-121, a 121-layer CNN, employs dense connections between layers to improve feature propagation and reuse, with 8 million parameters. Both architectures are pre-trained on ImageNet.

\subsection{Image preprocessing}
We resize the input images to maintain their aspect ratio while adjusting the shorter side to 232 or 256 pixels for the ResNet-50 and DenseNet-121, respectively. The images were then normalized using the mean and standard deviation values from the ImageNet dataset. Finally, the images are cropped to a uniform size of 224×224 pixels. To further enhance the performance of the model, we employed several data augmentation techniques \cite{cui18large, cubuk19autoaugment}. Specifically, we apply transformations such as rotation, mirroring, and shearing. 

\subsection{Evaluation metric}
Tool detection performance is assessed using the mean Average Precision (mAP) for multi-label classification tasks \cite{twinanda17endonet}. This metric is computed as the mean of the Average Precision (AP) over all classes and is defined as follows:

\begin{equation}
     mAP = \frac{1}{N_C} \sum_{t=1}^{N_C} \text{AP}_t
\end{equation}

where $N_C$ is the number of classes and $\text{AP}_t$ is the Average Precision for class $t$.

Average Precision (AP) is calculated by integrating the precision-recall curve, which represents the trade-off between precision and recall at different threshold levels. For a given class $t$, the Average Precision $AP_t$ is computed as:

\begin{equation}
AP_t = \sum_{n} (R_n - R_{n-1}) P_n
\end{equation}

where $P_n$ denotes precision and $R_n$ denotes the recall at the $n$-th threshold, both varying with the confidence threshold. 

\subsection{Experimental implementation and training setup}
The implementation of our approach was carried out using the PyTorch 2.0 framework. All experiments were conducted on a high-performance workstation equipped with an AMD Ryzen Threadripper PRO 3995WX 2.7GHz processor, 256 GB RAM, and an NVIDIA GeForce RTX A6000 GPU. The task of surgical tool presence detection is formulated as $N_C = 7$ binary classification tasks, corresponding to the number of tools in the Cholec80 dataset. The network training process utilized the Adam optimizer. We evaluated two initial learning rates: $1\times10^{-4}$ and $5\times10^{-4}$. The remaining hyperparameters were set as follows: a batch size of 64 and a maximum of 50 training epochs. Early stopping with a patience interval of 5 epochs was also incorporated. 

For this multi-label classification problem, the binary cross-entropy loss function was utilized for each tool. The total loss for a batch of $N_B$ images and $N_C$ classes is given by:

\begin{equation}
\text{Loss} = \frac{1}{N_B \times N_C} \sum_{n=1}^{N_B} \sum_{c=1}^{N_C} \left[ -y_{n,c} \log(\sigma(x_{n,c})) - (1 - y_{n,c}) \log(1 - \sigma(x_{n,c})) \right]
\end{equation}

where $y_{n,c}$ and $x_{n,c}$ are the target value and logit for the $n$-th sample and $c$-th class, respectively, and $\sigma(x)$ denotes the sigmoid function. Since frame-dependent information is not utilized (e.g., temporal relationships), all our experiments are conducted in an online mode, meaning that predictions are made without relying on future information.

\subsection{Adaptive fine-tuning implementation variations}
We implemented six versions of our proposed fine-tuning strategy (GFz) for surgical tool detection, corresponding to differences in the group of layers frozen at each freezing step, the method of updating the learning rate at each step, and network architecture:

\begin{itemize}
\item GFz-L : A percentage of the non-frozen layers in the model architecture are chosen to be frozen at each freezing step.
\item GFz-B1 : The model is decomposed into functional blocks. One block is frozen at each freezing step. 
\item GFz-B2 : The model is decomposed into functional blocks. One block is frozen at each freezing step. In contrast to the previous versions that update the learning rates based on Eq.~\ref{eq.2}, the learning rate weight for all model layers $\lambda_i$ is computed as the average of the learning rate weights of the corresponding block:
\begin{equation}
\alpha_i = \frac{\sum_{i \in B_j} \alpha_i}{|B_j|} \quad \forall i \in B_j
\end{equation}
\end{itemize}

These strategies are implemented for both the ResNet-50 and DenseNet-121 architectures. For ResNet-50, six functional blocks (residual blocks and classifier) are considered, while for DenseNet-121, nine functional blocks (dense blocks and classifier) are considered.

\section{Results}
\label{sec:5}

\begin{figure}[tb]	
\centering
\includegraphics[scale=0.5, width=0.8\linewidth]{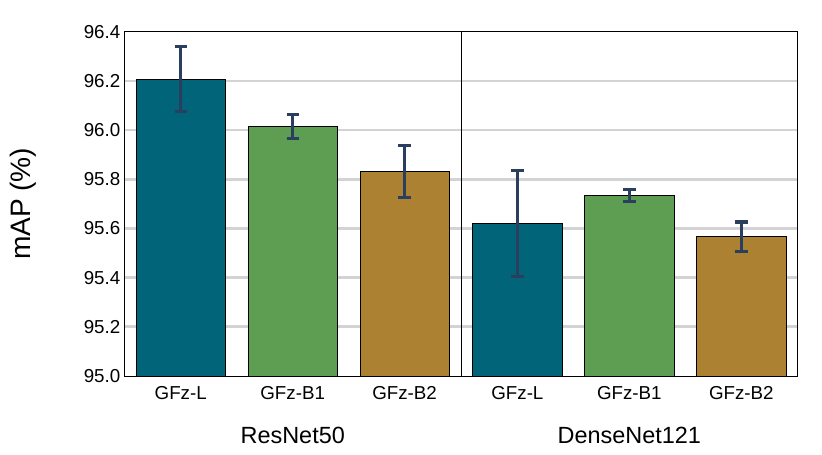}
\caption{Mean Average Precision (mAP) for all tools under the six implementations of the proposed fine-tuning strategy. \label{fig:3}}
\end{figure}  

\subsection{Surgical tool presence detection results}

The mean average precision (mAP) of all versions implemented for tool detection in the Cholec80 dataset, using the L40 split as proposed in \cite{twinanda17endonet}, are presented in Fig.~\ref{fig:3}. We independently calculate the Average Precision (AP) for each tool and then compute the mean AP across the seven tools. The experiments were repeated three times with different random seeds. Fig.~\ref{fig:3} illustrates the average mAP values along with the standard error in these three runs. All configurations demonstrated robust predictive performance, with mAP values exceeding 95\%. In particular, ResNet-based implementations demonstrated superior performance over DenseNet architectures. The highest performance was achieved with the ResNet50 architecture using a layer-based freezing strategy (GFz-L), in which 40\% of the non-frozen layers with the lowest importance index were frozen at each freezing step.

\subsection{Comparison with state-of-the-art methods}
From the previous section, we select the best performing configuration and compare its results with state-of-the-art approaches. As detailed in Section 4.1, prior studies have utilized different splits of the dataset to report metric results. To ensure a fair evaluation, we assessed the performance of our approach using the L40 and L30 splits and summarize these results in Table~\ref{tab:4} and Table~\ref{tab:5}.

\begin{table}[H]
\centering
\caption{Average precision of tool presence detection for the L40 split (32 videos for training and 40 videos for evaluation) and comparison with state-of-the-art approaches. The top scores for each tool are highlighted in bold.}\label{tab:4}
\renewcommand{\arraystretch}{1.4}  
\scalebox{0.9}{
\begin{tabular}{p{1.8cm} p{1.2cm} p{1.2cm} p{1.2cm} p{1.2cm} p{1.2cm} p{1.6cm} p{1.2cm} p{1.2cm}}
\headrow
\bf{Tool} & \bf{EndoNet \cite{twinanda17endonet}} & \bf{Endo3D \cite{chen18endo3d}} & \bf{GCN \cite{wang19graph}} & \bf{MTRC Net-CL \cite{jin20multitask}} & \bf{ResNet-LC-LV \cite{abdulbaki21deep}} & \bf{CNN-SE-MSF-LSTM \cite{jalal23laparoscopic}} & \bf{LAST \cite{tao23last}} & \bf{Our model} \\ 
Grasper         & 84.8  & 71.32 & -     & 84.7  & 87.40 & 91.0  & -     & \textbf{92.10} \\ 
Bipolar         & 86.9  & 69.72 & -     & 90.1  & 95.95 & \textbf{97.3} & -     & 95.98 \\ 
Hook            & 95.6  & 87.81 & -     & 95.6  & 99.45 & \textbf{99.8} & -     & 99.60 \\ 
Scissor         & 58.6  & 87.33 & -     & 86.7  & 92.73 & 90.3  & -     & \textbf{95.09} \\ 
Clipper         & 80.1  & 95.12 & -     & 89.8  & 98.50 & 97.4  & -     & \textbf{98.56} \\ 
Irrigator       & 74.4  & 96.43 & -     & 88.2  & 91.37 & 95.6  & -     & \textbf{97.05} \\ 
Specimen bag    & 86.8  & 94.97 & -     & 88.9  & 96.58 & \textbf{98.3} & -     & 96.61 \\ 
Mean            & 81.02 & 86.1  & 90.13 & 89.1  & 94.57 & 95.6  & 95.15 & \textbf{96.43} \\ 
\hline
\end{tabular}}
\end{table}

\begin{figure}[tb]	
\centering
\includegraphics[scale=0.8, width=0.82\linewidth]{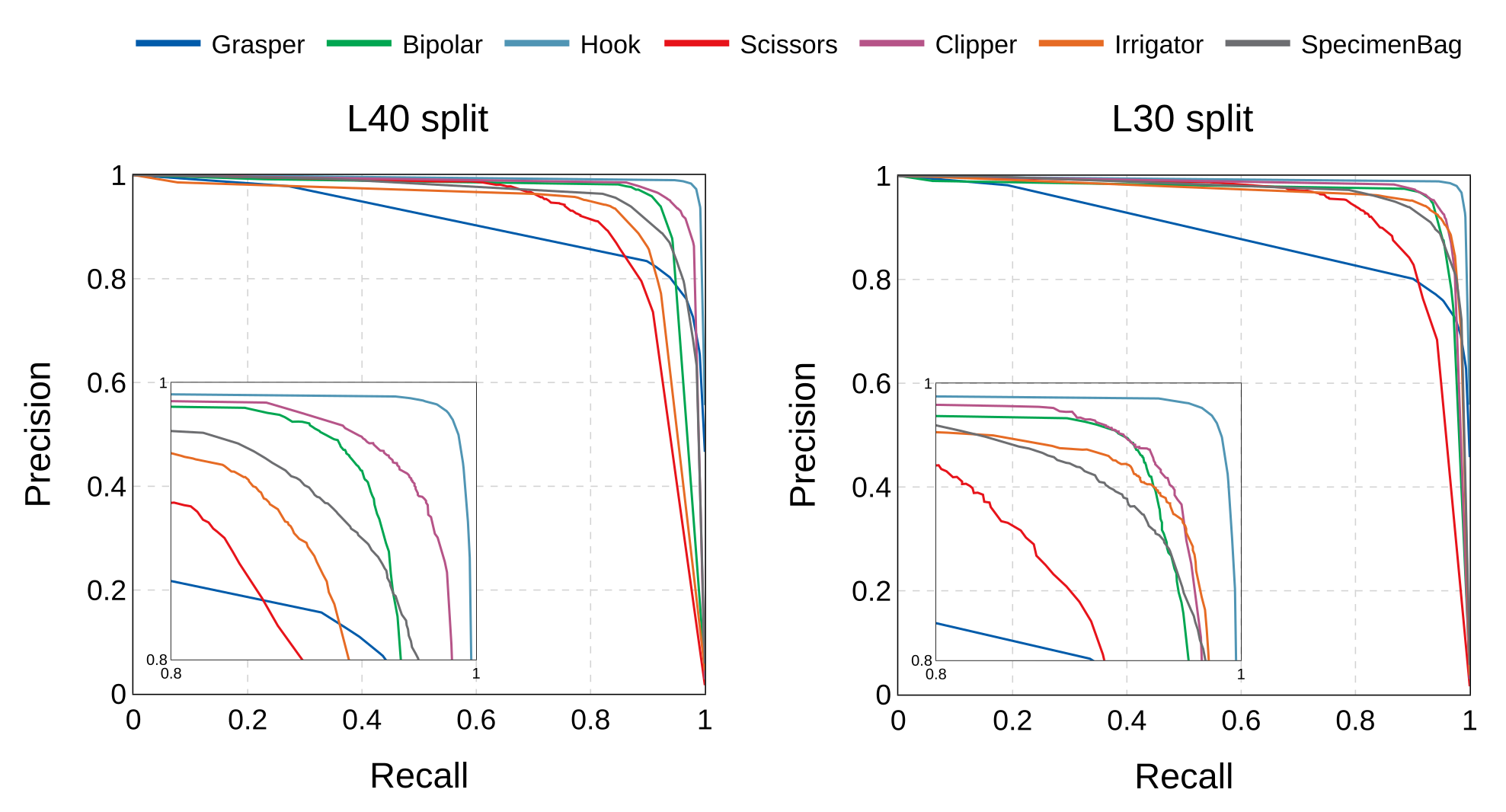}
\caption{Precision-Recall curves for the seven surgical tools using the L40 and L30 splits.  \label{fig:4}}
\end{figure}

Although following a simple approach, the proposed gradual freezing strategy significantly improved tool classification performance over previous approaches, which often involve high complexity and computational cost. The mean Average Precision (mAP) for all tools consistently demonstrated high detection performance regardless of the split employed, achieving 96.43\% for the L40 split and 96.32\% for the L30 split. In contrast with previous approaches, the high imbalance in the dataset did not seem to affect our results. Notably, the Scissor tool, which has the lowest number of appearances and exhibited low prediction performance in previous studies, showed better performance in our study. However, tools with a large number of appearances, such as Hook and Grasper, had contradictory results. The Hook tool achieved over 99\% mAP, while the Grasper tool had the lowest mAP, averaging 91\%. This discrepancy appears to highlight the sensitivity of our approach to multitool occurrences, as the simultaneous appearance of Grasper and Hook in the same frame is the most common occurrence in the Cholec80 dataset \cite{kondo21lapformer}. The Precision-Recall curves for the L40 and L30 splits, shown in Fig.~\ref{fig:4}, reflect these trends, with Grasper and Hook showing contradictory results. In general, the proposed model achieved the best prediction performance for most independent tools.

\begin{table}[tb]
\centering
\caption{Average precision of tool presence detection for the L30 split (40 videos for training and 30 videos for evaluation) and comparison with state-of-the-art approaches. The top scores for each tool are highlighted in bold.}\label{tab:5}
\renewcommand{\arraystretch}{1.4}  
\scalebox{0.95}{
\begin{tabular}{p{1.8cm} p{1.8cm} p{1.6cm} p{1.6cm} p{1.6cm} p{1.6cm}}
\headrow
\bf{Tool} & \bf{Vardazaryan \cite{vardazaryan18weakly}} & \bf{Nwoye \cite{nwoye19weakly}} & \bf{ResNet-LC-LV \cite{abdulbaki21deep}} & \bf{Jaafari\footnotemark[1] \cite{jaafari21towards}} & \bf{Our model} \\ 
Grasper         & 96.8  & \textbf{99.7} & 85.54 & 96.58 & 90.79 \\ 
Bipolar         & 94.2  & 95.6  & \textbf{95.88} & 95.04 & 95.73 \\ 
Hook            & 99.6  & \textbf{99.8} & 99.36 & 99.68 & 99.64 \\ 
Scissor         & 49.8  & 86.9  & 92.39 & 81.24 & \textbf{94.83} \\ 
Clipper         & 83.0  & 97.5  & \textbf{98.72} & 95.11 & 98.20 \\ 
Irrigator       & 93.3  & 74.7  & 95.89 & 93.71 & \textbf{97.97} \\ 
Specimen bag    & 94.0  & 96.1  & 96.83 & 94.92 & \textbf{97.04} \\ 
Mean            & 87.2  & 92.9  & 94.95 & 93.75 & \textbf{96.32} \\ 
\hline
\multicolumn{6}{l}{\footnotesize\footnotemark[1]Used the L20 split (60 videos for training and 20 videos for evaluation)} \\
\end{tabular}}
\end{table}

\begin{figure}[tb]	
\centering
\includegraphics[scale=1.0, width=0.95\linewidth]{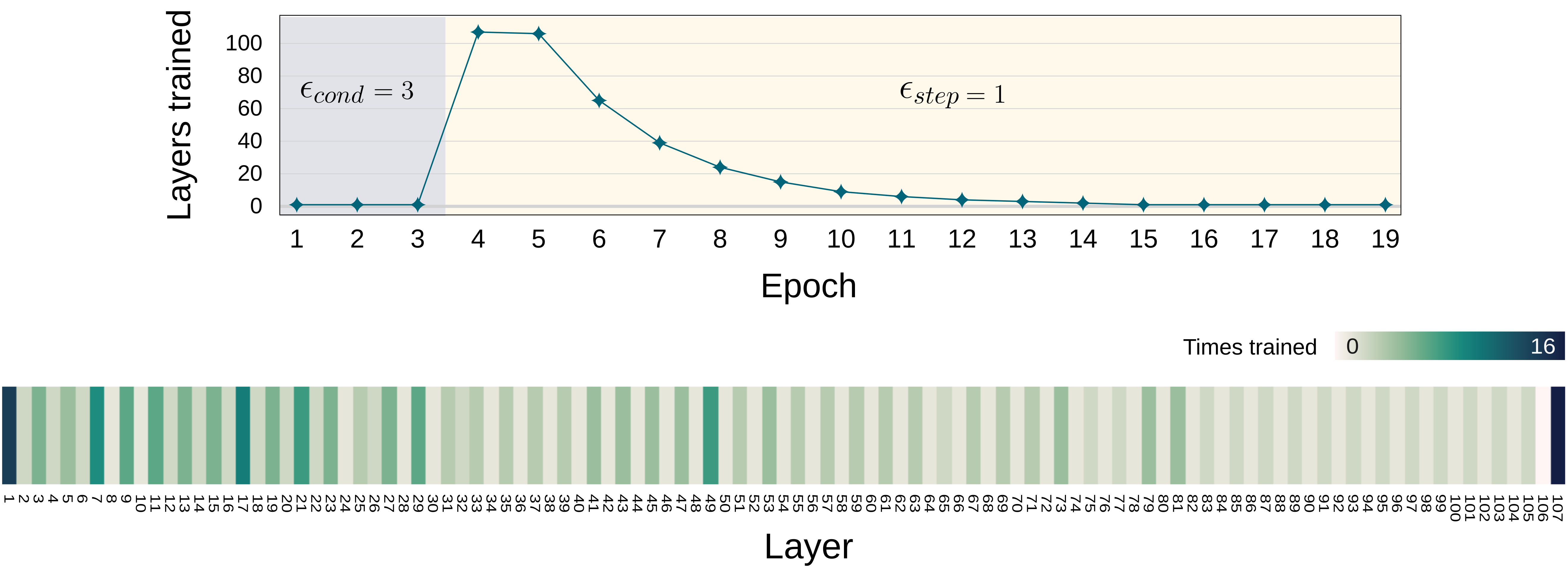}
\caption{\textbf{Top:} Number of layers trained across training epochs. \textbf{Bottom:} Heatmap representing the number of times each layer is trained across model layers.  \label{fig:5}}
\end{figure}  

Figure~\ref{fig:5} provides insight into the dynamics of our gradual freezing strategy during a representative training run in L30 split. The top graph shows the number of active (unfrozen) layers decreasing at each epoch. During the initial pre-conditioning phase ($\epsilon_{\text{cond}}=3$), only the final classifier layer is trained. Subsequently, after a brief full network fine-tuning phase, the number of trainable layers is progressively reduced by 40\% at each freezing step ($\epsilon_{\text{step}}=1$) until early stopping stops the process at Epoch 19.

The heatmap (bottom) reveals which parts of the network were trained most extensively. The training distribution is concentrated at the two extremes of the network: the initial feature extraction layers and the final, fully-connected classifier layer. This dual focus is expected. The early layers must adapt their general-purpose features from the source domain to the specific low-level visual characteristics of the surgical environment. Concurrently, the final classifier layer, being randomly initialized, requires extensive training to learn the new surgical tool classification task. In contrast, the intermediate layers, which learn more abstract features, were frozen much earlier in the process. This highlights the method's ability to efficiently preserve stable, mid-level features while selectively adapting the most critical parts of the network to the new domain.

\subsection{Comparison with other fine-tuning strategies}
The proposed fine-tuning strategy was compared with both conventional and recent approaches.
We examine eight distinct fine-tuning strategies used for adapting pre-trained models to the target datasets:

\begin{itemize}
    \item \textbf{Full Fine-tuning (FT)}: A conventional fine-tuning strategy where all layers of the pre-trained model are unfrozen and trained on the target dataset \cite{hinton06reducing}.
    \item \textbf{Linear Probing (LP)} \cite{kumar22fine}: An approach where all pre-trained model layers are frozen except for the final classifier layer. Only this last layer is trained, which learns a linear mapping from the fixed, pre-trained features to the target classes.
    \item \textbf{Gradual Unfreeze (Last → First) (G-LF)} \cite{howard18universal}: In this strategy, the model is conceptually divided into sequential blocks of layers. Initially, all blocks are frozen except for the final block (e.g., the classifier), which is trained. After a set number of epochs, the preceding block is also unfrozen, and the process continues sequentially from the last block towards the first. The rationale is to adapt the more task-specific blocks before fine-tuning the more general feature extraction blocks.
    \item \textbf{Gradual Unfreeze (First → Last) (G-FL)} \cite{mukherjee20distilling}: This method also divides the model into blocks but reverses the unfreezing order. Training begins with only the first block of layers unfrozen. Subsequently, the next block in the sequence is unfrozen, and this process repeats from the first block towards the last. The underlying idea is to allow for a sequential adaptation of the model, starting from general low-level feature blocks and moving towards more specific high-level ones.
    \item \textbf{Gradual Unfreeze (Last/All) (LP-FT)}\cite{kumar22fine}: A two-stage approach that first trains the model using Linear Probing for a set number of epochs. Afterwards, all layers are unfrozen for a full fine-tuning phase.
    \item \textbf{$L^1$-\textit{SP} Regularization} \cite{li18explicit}: These methods add a regularization term to the loss function that penalizes the difference between the fine-tuned weights ($\omega_t$) and the original pre-trained weights ($\omega_s^0$). This encourages the model to maintain weights close to the pre-trained initialization to counter catastrophic forgetting. We test both $L^1$-norm and $L^2$-norm variants, defined as:
        \begin{equation}
        \label{eq:8}
        \gamma^{\ell 1}(\omega) = a {|| \omega_{t} - \omega^{0}_{s} ||}_{1} + b {|| \hat{\omega}_{t} ||}_{1} ,
        \end{equation}
    \item \textbf{$L^2$-\textit{SP} Regularization} \cite{li18explicit}: Similar to $L^1$-SP, but uses an $L^2$-norm based regularization ($\gamma^{\ell 2}(\omega)$), penalizing larger weights more heavily, defined by
        \begin{equation}
        \label{eq:10}
        \gamma^{\ell 2}(\omega) = a || \omega_{s} - \omega^{0}_{s} ||^2_{2} + b || \hat{\omega}_{t} ||^2_{2}
        \end{equation}
    \item \textbf{Auto-RGN} \cite{lee23surgical}: A fine-tuning method that adjusts the learning rate for each layer based on its Relative Gradient Norm (RGN), as defined in Eq.~\ref{eq.1}. The premise is that layers with a higher RGN are contributing more to the learning process for the new task and should be updated with a higher learning rate.
    \item \textbf{Low-Rank Adaptation (LoRA)} \cite{hu21lora}: A parameter-efficient fine-tuning method where the original pre-trained weights are frozen. Small, trainable low-rank matrices are added to certain layers, and only these new matrices are updated during training. This significantly reduces the number of trainable parameters. We adapt this technique, typically used for transformers, to the CNN layers.
    \item \textbf{Gradual Freezing (GFz)}: Our proposed adaptive fine-tuning strategy based on gradual freezing.
\end{itemize}

Table~\ref{tab:6} summarizes the mAP results for ResNet50 and DenseNet121 architectures using the conventional fine-tuning approach (Full FT) and the proposed fine-tuning approaches. The proposed approach improves prediction performance for both architectures. ResNet50 benefits the most from the GFz strategy, increasing mAP by 0.84\%, while DenseNet121 also shows improvement with an mAP increase of 0.7\%.

\begin{table}[tb]
\centering
\caption{Mean Average Precision (mAP) for conventional full fine-tuning (Full FT) and proposed gradual freezing based fine-tuning strategies (GFz) for ResNet50 and DenseNet121 architectures.}\label{tab:6}
\renewcommand{\arraystretch}{1.2}  
\scalebox{0.95}{
\begin{tabular}{p{4cm} p{2.5cm} p{2.5cm}}
\headrow
\bf{FT Strategy} & \bf{ResNet50} & \bf{DenseNet121} \\ 
Full FT              & $95.41 \pm 0.15$ & $95.06 \pm 0.20$ \\ 
Proposed FT (GFz-L)  & $96.21 \pm 0.13$ & $95.62 \pm 0.22$ \\
Proposed FT (GFz-B1) & $96.01 \pm 0.05$ & $95.73 \pm 0.02$ \\
Proposed FT (GFz-B2) & $95.83 \pm 0.11$ & $95.57 \pm 0.05$ \\
\hline
\end{tabular}}
\end{table}

\begin{figure}[bt]	
\centering
\includegraphics[scale=0.5, width=0.85\linewidth]{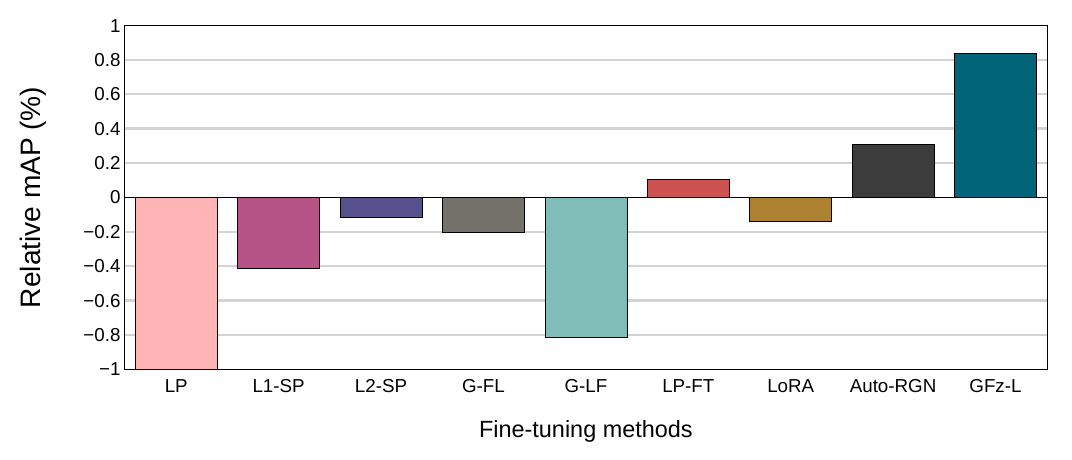}
\caption{Relative mean Average-Precision for a ResNet50 architecture across various fine-tuning strategies. \label{fig:6}}
\end{figure}  

\begin{figure}[tb]	
\centering
\includegraphics[scale=1.0, width=0.85\linewidth]{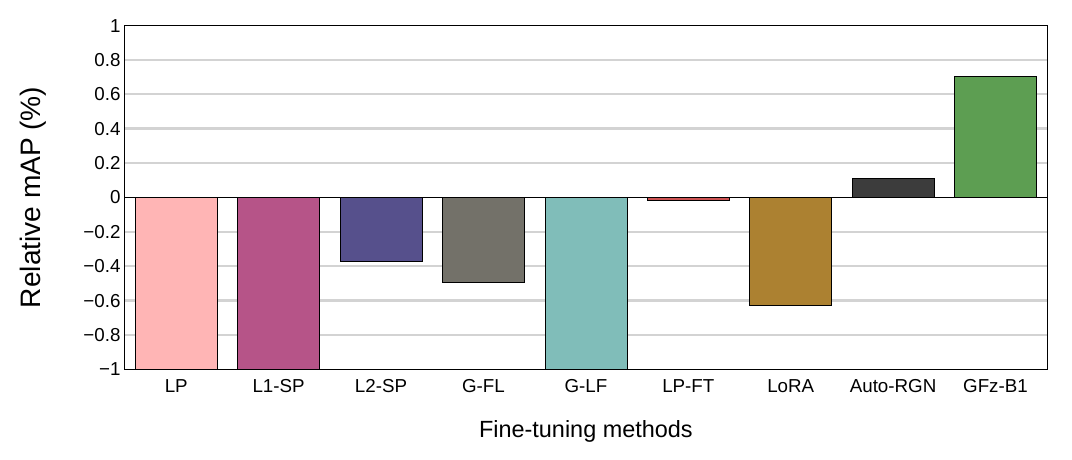}
\caption{Relative mean Average-Precision for a DenseNet121 architecture across various fine-tuning strategies. \label{fig:7}}
\end{figure}  

Full fine-tuning has traditionally been the default approach for transfer learning in most studies. Consequently, we evaluated the performance of various fine-tuning strategies based on the relative mean average precision ($\overline{\text{mAP}}$) in comparison to the Full fine-tuning approach, as given by:
    \begin{equation}
        \overline{\text{mAP}}_{T} = \frac{\text{mAP}_{T} - \text{mAP}_{Full}}{\text{mAP}_{Full}} 
    \end{equation}
where $\text{mAP}_{T}$ is the mAP for the fine-tuning strategy $T$, and $\text{mAP}_{Full}$ is the mAP for the Full fine-tuning strategy. The results are depicted in Fig.~\ref{fig:6} and Fig.~\ref{fig:7} for the ResNet50 and DenseNet121 architectures, respectively.

For ResNet50, the GFz-L approach achieves an improvement of over 0.8\% compared to Full Fine-Tuning. Among other fine-tuning strategies, only Auto-RGN and LP+FT show enhanced performance. Gradually unfreezing, which does the opposite of our approach, is significantly affected by the unfreezing direction. Unfreezing from the first layer, which aligns with our method's focus on training bottom layers, yields performance close to Full FT. Conversely, unfreezing from the top layers results in the second-worst performance, surpassed only by LP. LP, which trains solely the classifier layer, appears inadequate for adapting to the target dataset due to the substantial difference between source and target domains.

DenseNet121 exhibits a similar trend, with GFz-B1 showing an improvement of approximately 0.7\%. Auto-RGN also yields a slight performance increase, while all other fine-tuning approaches result in decreased performance compared to Full FT. For both architectures, most fine-tuning methods, including regularization-based, LoRA, and gradual freezing, lead to performance drops relative to the baseline full fine-tuning.

\subsection{Ablation studies}

We conducted several ablation studies to determine the impact of different parameters on the performance of the proposed approach. Specifically, we analyzed the effects of the number of epochs for pre-conditioning, the freezing frequency, and the layer grouping method. For these experiments, we focused solely on ResNet-50 variations, as they provided the best results, as demonstrated in the previous section. Each experiment was repeated three times with different random seeds, and we report the mean and standard deviation for the performance metrics.

\subsubsection{Number of epochs for pre-conditioning $\epsilon_{\text{cond}}$}
\begin{figure}[tb]	
\centering
\includegraphics[width=0.6\linewidth]{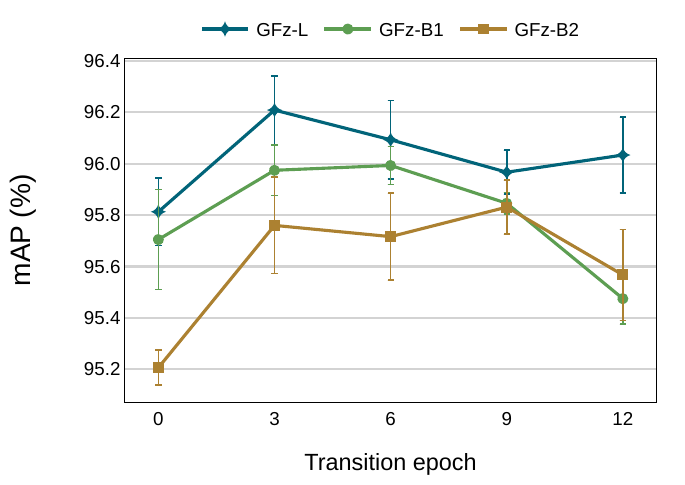}
\caption{Mean Average Precision (mAP) performance across different numbers of pre-conditioning epochs ($\epsilon_{\text{cond}}$) for layer-wise (GFz-L) and block-wise (GFz-B1, GFz-B2) Gradual Freezing methods. \label{fig:9}}
\end{figure}

Pre-conditioning is achieved by freezing all pre-trained model weights except for the last classification layer and training this single output layer for $\epsilon_{\text{cond}}$ epochs. We evaluated the effect of this parameter by considering values between 0 and 12 training iterations before starting the gradual freezing phase. Figure~\ref{fig:9} illustrates the mAP performance for the proposed ResNet-based variations.

\begin{figure}[tb]	
\centering
\includegraphics[width=0.57\linewidth]{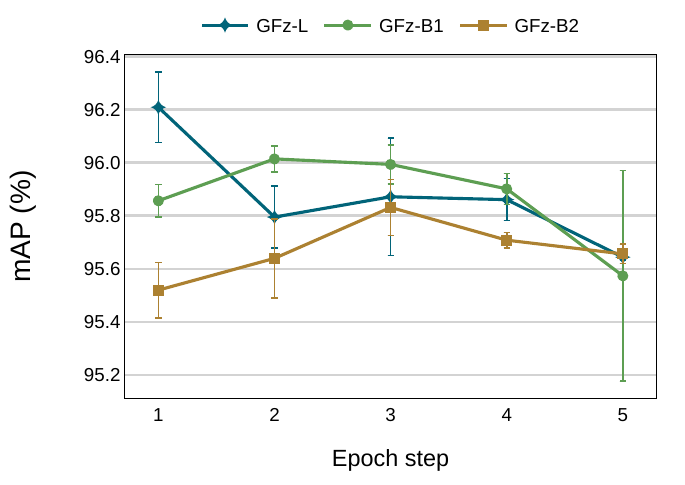}
\caption{Mean Average Precision (mAP) performance across different freezing frequencies. \label{fig:10}}
\end{figure}  

\begin{figure}[tb]	
\centering
\includegraphics[width=0.56\linewidth]{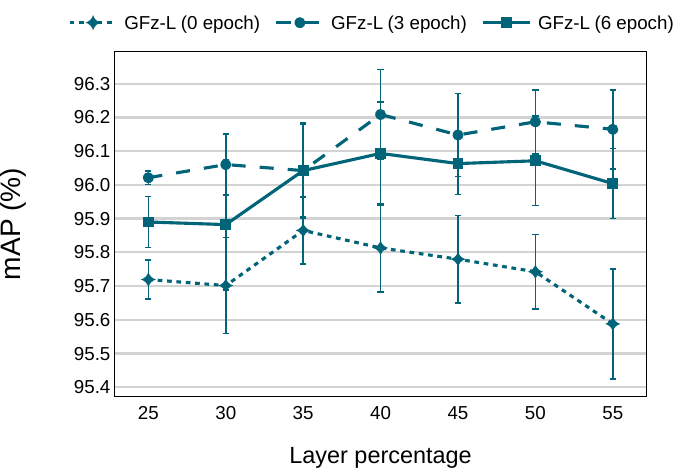}
\caption{Mean Average Precision (mAP) performance across different percentages of non-frozen layers selected for freezing, with the optimal performance observed at 40\% when $\epsilon_{\text{cond}} = 3$. \label{fig:11}}
\end{figure}  

The results indicate that no pre-conditioning leads to low performance, while optimal performance is achieved with values between 3 and 9 epochs. Additionally, freezing based on layer groups rather than functional blocks achieves higher performance, with the best results observed at $\epsilon_{\text{cond}} = 3$.

\subsubsection{Freezing frequency $\epsilon_{\text{step}}$}
We evaluated the effect of the freezing frequency, defined as the number of $\epsilon_{\text{step}}$ epochs for updating the importance index and freezing the lowest group of layers based on this index. We restricted the values for the freezing frequency to be less than or equal to the patience defined for the early stop ($\epsilon_{\text{step}} \leq 5$). This ensures that the early stop does not halt training before the next freezing step. Figure~\ref{fig:10} shows the mAP performance for the ResNet variations.

The results indicate that top performance is achieved with smaller $\epsilon_{\text{step}}$ values. The best performance is observed for the case of grouping based on layers (GFz-L), achieving optimal results with a single epoch, meaning that a new group of layers is frozen at every training iteration.

Based on these empirical results, the optimal hyperparameters ($\epsilon_{\text{cond}}=3$, $\epsilon_{\text{step}}=1$, and a 40\% freezing ratio for GFz-L) were selected for our final model comparisons. This empirical tuning via ablation follows standard practice for optimizing the performance of deep learning models.

\subsubsection{Percentage of freezing layers}
During the gradual freezing stage, a group of layers is frozen at each freezing step. This group can range from a single layer to an entire block containing all model layers. In the GFz-L version, layers are grouped based on a predefined percentage of the total non-frozen layers in the model at each freezing step. We evaluated the effect of various predefined percentages, along with pre-conditioning ($\epsilon_{\text{cond}}$) epochs varying between 0 and 6.

The results are shown in Figure~\ref{fig:11}. Although performance remains similar in all evaluated cases, it can be observed that the best mAP performance is achieved with $\epsilon_{\text{cond}} = 3$ and selecting 40\% of the non-frozen layers to be frozen at each freezing step.

\subsection{Generalizability on the CATARACTS Dataset}
To validate that the benefits of our GFz approach are not limited to a single surgical domain, we evaluated its performance on the CATARACTS dataset \cite{alhajj21cataracts}. The primary goal of this experiment was not to establish a new state-of-the-art benchmark for this specific dataset, but rather to assess the generalizability of our fine-tuning strategy on a distinct surgical domain. This provides a demanding out-of-domain validation, confirming that the benefits of GFz are based in a more fundamental principle of adaptive learning. We used the same ResNet-50 architecture and first compared our best-performing GFz-L strategy against the conventional full fine-tuning (Full FT) baseline. As summarized in Table~\ref{tab:7}, our GFz-L method achieved a mean Area Under the Curve (AUC) of $96.78\%$, a significant improvement of approximately $2.4\%$ over the $94.43\%$ achieved by Full FT.

\begin{figure}[tb]	
\centering
\includegraphics[width=0.9\linewidth]{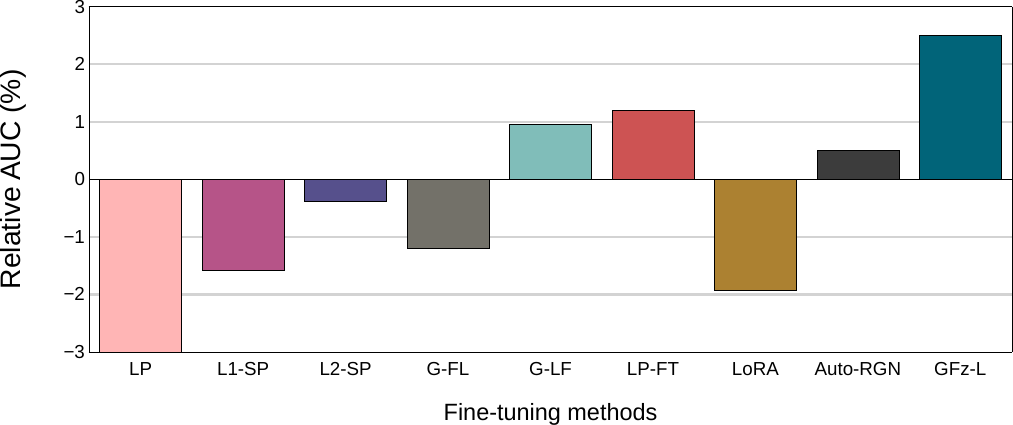}
\caption{Relative mean AUC for various fine-tuning strategies on the CATARACTS dataset using a ResNet-50 architecture. Performance is shown relative to the Full Fine-tuning (FT) baseline. \label{fig:8}}
\end{figure}  

\begin{table}[tb]
\centering
\caption{Mean Area Under the Curve (AUC) on the CATARACTS dataset. Comparison between conventional full fine-tuning (Full FT) and our proposed layer-based gradual freezing (GFz-L) strategy using a ResNet-50 architecture.}\label{tab:7}
\renewcommand{\arraystretch}{1.2}  
\scalebox{0.95}{
\begin{tabular}{p{4cm} p{2.5cm}}
\headrow
\bf{FT Strategy} & \bf{ResNet50} \\
Full FT              & $94.43 \pm 0.54$ \\ 
Proposed FT (GFz-L)  & $96.78 \pm 0.16$ \\
\hline
\end{tabular}}
\end{table}

To provide a more comprehensive benchmark, we also compared our approach against several other established fine-tuning strategies. Figure~\ref{fig:8} illustrates the performance of these methods relative to the Full FT baseline. The comparison reveals that several common techniques, including Linear Probing (LP), L1-SP regularization, and LoRA, resulted in a degradation in performance on this dataset. While some methods like Gradual Unfreezing (G-LF) and Auto-RGN offered marginal improvements, our proposed GFz-L strategy substantially outperformed all other tested methods, yielding the highest relative performance gain. This results provides strong evidence for the generalizability and effectiveness of the proposed gradual freezing strategy.

\subsection{Computational efficiency and real-time applicability}
For real-time applicability in surgical environments, inference speed is a critical metric. Since our gradual freezing method modifies the training procedure but results in a final model architecture identical to a conventionally fine-tuned one, their inference speeds are comparable. We quantified this on an NVIDIA GeForce RTX A6000 GPU, measuring an average inference time of approximately 1.15 ms per frame. This translates to a throughput of about 871 frames per second (fps), which significantly exceeds the 25-30 fps typically required for real-time surgical video analysis, confirming the suitability of the model for deployment.

The primary efficiency gains of our GFz strategy are realized during the training process. This improvement comes from the progressive reduction in trainable parameters as the layers are frozen. As illustrated in Figure~\ref{fig:5}, in the final stages of training, only a small fraction of the most important layers remain active. This dynamic reduction of the network's trainable complexity decreases the computational load for the backward pass. To quantify this, we compared our best performing GFz-L model (ResNet-50 on Cholec80) against a standard full fine-tuning baseline on the same hardware. The complete fine-tuning took 101 minutes to complete, while our GFz approach was completed in 81 minutes, a 20\% reduction in training time. This was accompanied by a decrease in the average GPU memory consumption from 5.78 GB to 4.73 GB, since fewer gradients and optimizer states need to be stored.

\section{Discussion}
\label{sec:6}

\subsection{Comparison of Gradual Freezing with Other Fine-Tuning Methods}

The field of transfer learning includes a variety of fine-tuning strategies. To position the contribution of our Gradual Freezing (GFz) method, it is useful to compare it with other major approaches. Our method is a dynamic learning strategy, in contrast to both static fine-tuning rules and computationally intensive search-based methods.

Static strategies, such as full fine-tuning, linear probing, or freezing a fixed block of layers, apply a single, time-invariant rule for model plasticity. Although simple, these methods can be suboptimal, as they do not adapt to the learning dynamics of the model. Search-based strategies, such as those that use metaheuristic algorithms to find an optimal set of layers to prune, treat the problem as a combinatorial search for a single optimal static architecture. This can be highly effective, but often comes at a prohibitive computational cost.

Our GFz method occupies a middle ground. It is dynamic, assuming the optimal model configuration changes over time. It is also principled, using gradient-based signals from the learning process to guide decisions, and efficient, operating within a single training pass. Table~\ref{tab:8} provides a comparative summary of these approaches.

\begin{table}[tb]
\centering
\caption{A comparative analysis of fine-tuning strategies.}\label{tab:8}
\renewcommand{\arraystretch}{1.4} 
\scalebox{0.85}{ 
\begin{tabular}{p{2.5cm} p{2.4cm} p{2.8cm} p{3.5cm} p{3.5cm}}
\headrow
\textbf{Method} & \textbf{Methodological Paradigm} & \textbf{Adaptivity Mechanism} & \textbf{Key Advantage} & \textbf{Key Limitation} \\
\hline
Full Fine-Tuning (FT) & Static Transfer & Full backpropagation on all layers & Maximum model plasticity; high potential performance. & High risk of catastrophic forgetting and overfitting. \\

Linear Probing (LP) & Static Transfer & Trains only the final classifier layer & Excellent feature preservation; computationally cheap. & Extremely low plasticity; often underfits. \\

Static Partial Freezing & Static Transfer & Heuristic freezing of a fixed layer block & Balances plasticity and preservation; simple to implement. & The choice of layers to freeze is non-adaptive and often suboptimal. \\

Gradual Unfreezing & Scheduled Learning & Time-based, sequential unfreezing of layers & Provides a learning curriculum from simple to complex. & Schedule is a fixed hyperparameter; may adapt too slowly. \\

Regularization-Based (L2-SP) & Regularized Learning & Adds a penalty for deviating from pre-trained weights & Explicitly combats catastrophic forgetting across all layers. & Adds a sensitive regularization hyperparameter to tune. \\

Parameter-Efficient (LoRA) & Adapter-based Learning & Adds small, low-rank trainable matrices & Drastically reduces trainable parameters; memory efficient. & May have limited capacity to capture complex adaptations. \\

Metaheuristic Fine-Tuning (GA/PSO) & Combinatorial Search & Population-based search for an optimal static configuration & Can discover a highly optimized, sparse architecture for the task. & Extremely high computational search cost; inflexible once training starts. \\

\textbf{Proposed Gradual Freezing (GFz)} & Dynamic Learning Strategy & Gradient-based, sequential freezing of layers & Adapts to learning dynamics in a single pass; efficient and stable. & The freezing schedule and ratio are tunable hyperparameters. \\
\hline
\end{tabular}}
\end{table}

\subsection{Rationale for Gradual Freezing in surgical domains}

The GFz strategy excels due to its balanced approach in preserving pre-trained features while adapting to new tasks, essential for surgical video analysis which demands robust general features and specialized adaptability. The robustness of our method is directly evidenced by its high performance in the Cholec80 and CATARactS datasets, which are inherently noisy due to real-world surgical conditions such as instrument glare, smoke, fluid occlusions, and variable illumination. GFz excels in these environments because it inherently resists overfitting to such noise. By identifying and freezing layers that have stabilized (low RGN), our method preserves robust, general feature hierarchies learned from ImageNet. This gradual constraint forces the model to base its predictions on these more abstract and reliable features rather than memorizing superficial, noisy patterns from the target data, thus improving generalization.



The training dynamics illustrated in Figure~\ref{fig:5} provides valuable insight into the interpretability of the method. Our method automatically concentrates training on the initial `sensory' layers and the final `classification' layer. This is because the early layers must adapt to the low-level visual characteristics of the surgical domain, while the final layer learns the new task from scratch. By progressively solidifying the intermediate layers that represent more abstract knowledge, our approach efficiently preserves their learned feature representations, effectively locking in valuable knowledge.

Finally, the generalizability of the method was thoroughly tested on the CATARACTS dataset. This demanding out-of-domain test validated the strategy's effectiveness in managing the significant domain shift from general-purpose images (ImageNet) to a distinct surgical environment (ophthalmic surgery). Its success in a domain with different anatomy, lighting and instrumentation demonstrates that the principles of GFz are robust and not limited to a single task, providing strong evidence for the broad applicability of our adaptive fine-tuning strategy.


\subsection{Limitations and future work}
Although our method shows promising results, there are several limitations to consider. First, efficiency improvements are limited to the training phase and do not affect the inference time of the model. Secondly, the gradual freezing schedule depends on pre-set hyperparameters, such as $\epsilon_{\text{step}}$ and the percentage of layers to freeze, which may need adjustment for optimal performance on different datasets or architectures. Morover, our evaluations were conducted only on CNN-based architectures, and further investigation is needed to assess the method's applicability to other models like Vision Transformers. Lastly, our evaluation relies on demonstrating consistent performance gains across multiple architectures, datasets, and repeated trials. Although formal statistical hypothesis testing was not conducted, the consistency and generalizability of the observed improvements provide strong evidence for the effectiveness of our method. In future efforts, statistical validation could be integrated by increasing the number of training trials.

Future work could focus on developing a fully automated fine-tuning strategy where the freezing schedule itself is learned dynamically based on model state, removing the need for manual hyperparameter tuning. Furthermore, extending this adaptive fine-tuning approach to other complex medical imaging tasks, such as segmentation or multi-modal analysis, presents an exciting avenue for future research.

\section{Conclusion}
\label{sec:7}
This paper introduced a novel adaptive fine-tuning strategy, Gradual Freezing (GFz), to address the challenges of surgical tool presence detection. By systematically preserving pre-trained knowledge while adapting to the target domain, our method demonstrates a significant performance improvement over established fine-tuning techniques and state-of-the-art approaches on the Cholec80 laparoscopic dataset. Furthermore, we confirmed the generalizability of our strategy through successful application to the distinct CATARACTS ophthalmic surgery dataset.

The proposed approach is not only effective but also improves training efficiency by reducing the computational load of the backward pass as layers are incrementally frozen. The comprehensive ablation studies provided insight into the method's key parameters, confirming its robustness. These findings establish gradual freezing as a promising and practical strategy for developing more accurate and efficient models for tool recognition in diverse surgical environments. Future work, as detailed in the discussion, will aim to further automate and extend this approach to other complex medical imaging tasks.

\section*{acknowledgements}
This work was supported in part by the Japan Science and Technology Agency (JST) CREST including AIP Challenge Program under Grant JPMJCR20D5, and in part by the Japan Society for the Promotion of Science (JSPS) Grants-in-Aid for Scientific Research (KAKENHI) under Grant 25K21247.

\section*{conflict of interest}
The authors declare no conflicts of interest.


\printendnotes

\bibliography{biblio}



\end{document}